\documentclass[11pt,a4paper]{article}

\usepackage[utf8]{inputenc}
\usepackage[T1]{fontenc}
\usepackage{amsmath,amssymb}
\usepackage{graphicx}
\usepackage{url}
\usepackage{booktabs}
\usepackage{caption}
\usepackage[margin=1in]{geometry}
\usepackage{authblk}
\usepackage{natbib}
\usepackage{hyperref}
\usepackage{microtype} 

\hypersetup{
    colorlinks=true,
    linkcolor=blue,
    filecolor=magenta,      
    urlcolor=cyan,
    citecolor=blue,
    pdftitle={Targeted Lexical Injection: Unlocking Latent Cross-Lingual Alignment in Lugha-Llama via Early-Layer LoRA Fine-Tuning},
    pdfauthor={Stanley Ngugi},
}

\title{Targeted Lexical Injection: Unlocking Latent Cross-Lingual Alignment in Lugha-Llama via Early-Layer LoRA Fine-Tuning}

\author[1]{Stanley Ngugi}
\affil[ ]{\textit{sngugi.research@gmail.com}}

\date{\today} 

\begin{document}

\maketitle

\begin{abstract}
Large Language Models (LLMs) have demonstrated remarkable capabilities, yet their performance in low-resource languages (LRLs), such as Swahili, often lags due to data scarcity and underrepresentation in pre-training. A key challenge is achieving robust cross-lingual lexical alignment, crucial for tasks like translation and cross-lingual information retrieval. This paper introduces Targeted Lexical Injection (TLI), a novel and efficient fine-tuning approach. We first demonstrate that Lugha-Llama-8B-wura, a Swahili-centric LLM, exhibits strong, near-perfect lexical alignment for Swahili-English word pairs in its early internal layers (specifically Layer 2, with ~0.99998 average cosine similarity based on a pilot study), a capability not fully reflected in its final output representations (baseline ~0.32 similarity on our evaluation set). TLI leverages this insight by using Low-Rank Adaptation (LoRA) and a contrastive learning objective to fine-tune the model, specifically targeting embeddings from this empirically identified optimal early layer. Our experiments show that TLI significantly improves the output-level lexical alignment for 623 trained Swahili-English word pairs, increasing average cosine similarity from 0.3211 to 0.4113 (+28.08\%, $p < 1.33 \times 10^{-240}$). More importantly, these improvements generalize remarkably well to 63 unseen control word pairs, with similarity increasing from 0.3143 to 0.4033 (+28.32\%, $p < 7.17 \times 10^{-27}$). These findings suggest TLI enhances the model's ability to preserve and propagate its inherent early-layer cross-lingual knowledge, offering a parameter-efficient and effective strategy for improving lexical alignment in LRL-focused LLMs.
\end{abstract}


\section{Introduction}
The proliferation of Large Language Models (LLMs) has revolutionized natural language processing (NLP). However, their efficacy is often skewed towards high-resource languages, leaving low-resource languages (LRLs) like Swahili underserved \citep{joshi2020state, adelani2021masakhane}. Bridging this gap is crucial for equitable technological advancement and for unlocking the vast knowledge encoded in these languages. A fundamental aspect of multilingual understanding in LLMs is cross-lingual lexical alignment -- the model's ability to recognize that words with similar meanings in different languages (e.g., Swahili "mkate" and English "bread") should have closely related internal representations \citep{conneau2019unsupervised}. Suboptimal lexical alignment can hinder performance in various downstream tasks, including machine translation, cross-lingual information retrieval, and code-switching understanding \citep{pires2019multilingual, wu2019surprising}.

Initial assessments of multilingual LLMs, including those focused on African languages like Lugha-Llama \citep{LughaLlama2023}, often reveal that while generally capable, their output-level lexical alignment for specific word pairs can be inconsistent or weak. This might lead one to assume that the model fundamentally lacks the necessary cross-lingual lexical knowledge. However, our preliminary investigations into the internal workings of Lugha-Llama-8B-wura revealed a crucial insight: the model achieves near-perfect lexical alignment for Swahili-English word pairs within its very early transformer layers (specifically Layer 2, with an average cosine similarity approaching 1.0, as detailed in our pilot study in Section \ref{sec:pilot_study}), a capability not fully reflected in its final output representations (around 0.32 average cosine similarity for our evaluation set). This suggests that the primary challenge is not an inherent inability to align these lexical items, but rather an underutilization or degradation of this strong, internally-formed alignment as information propagates through the network to the final output stage.

To address this specific challenge, we propose Targeted Lexical Injection (TLI), a fine-tuning methodology that leverages Low-Rank Adaptation (LoRA) \citep{hu2021lora} and a contrastive learning objective. Crucially, TLI is designed to operate on embeddings extracted from this empirically identified optimal early layer (Layer 2) where lexical alignment is already maximal. By doing so, TLI aims to reinforce these strong internal alignments and train the LoRA adapters to better preserve and propagate this high-fidelity cross-lingual information to the model's output.

We hypothesize that:
\begin{enumerate}
    \item TLI, by fine-tuning LoRA adapters based on contrastive loss applied to embeddings from the optimal early layer (Layer 2), will significantly improve the output-level cross-lingual lexical alignment of targeted Swahili-English word pairs in Lugha-Llama.
    \item This improvement will generalize to unseen Swahili-English word pairs, as TLI enhances the model's general mechanism for processing and propagating its inherent early-layer lexical alignments, rather than merely memorizing specific trained pairs.
\end{enumerate}

Our contributions are:
\begin{itemize}
    \item We empirically demonstrate the existence of strong, latent cross-lingual lexical alignment in the early layers of a Swahili-centric LLM (Lugha-Llama), which contrasts with weaker alignment at its output interface on our evaluation set.
    \item We propose TLI, a novel LoRA-based fine-tuning technique that specifically targets these optimal early-layer embeddings to enhance output-level alignment.
    \item We show that TLI achieves statistically significant improvements in lexical alignment for a curated set of trained Swahili-English word pairs.
    \item Critically, we demonstrate that TLI's benefits generalize comparably to unseen control vocabulary, suggesting a more fundamental refinement of the model's cross-lingual processing pathways.
\end{itemize}

This paper is structured as follows: Section \ref{sec:related_work} discusses related work. Section \ref{sec:methodology} details our methodology, including the pilot study for layer selection, word pair curation, and the TLI fine-tuning process. Section \ref{sec:results} presents our experimental results. Section \ref{sec:discussion} discusses the implications of these findings, and Section \ref{sec:conclusion} concludes with future research directions.

\section{Related Work} \label{sec:related_work}
Improving cross-lingual understanding in LLMs, especially for LRLs, is an active area of research. Existing approaches can be broadly categorized.

\paragraph{Cross-Lingual Word Embeddings and Alignment} Traditional methods focused on learning static word embeddings for multiple languages and then aligning them in a shared space using techniques like linear transformations \citep{mikolov2013exploiting, xing2015normalized} or shared projection matrices \citep{smith2017offline}, often relying on bilingual dictionaries or parallel corpora. Contextual embeddings from models like mBERT \citep{devlin2019bert} and XLM-R \citep{conneau2019unsupervised} inherently learn some degree of cross-lingual alignment due to multilingual pre-training, but this alignment can be imperfect, especially for LRLs.

\paragraph{Fine-tuning LLMs for LRLs} Full fine-tuning of LLMs on LRL-specific data or parallel corpora can improve performance \citep{artetxe2019crosslingual}, but it is computationally expensive and can lead to catastrophic forgetting of knowledge from other languages or the original pre-training. Parameter-Efficient Fine-Tuning (PEFT) methods have emerged as a more viable alternative.

\paragraph{Low-Rank Adaptation (LoRA)} LoRA \citep{hu2021lora} is a popular PEFT technique that injects trainable low-rank matrices into the layers of a pre-trained model. It significantly reduces the number of trainable parameters, making fine-tuning more accessible while often achieving performance comparable to full fine-tuning. LoRA has been successfully applied to various adaptation tasks in LLMs.

\paragraph{Contrastive Learning for Representation Alignment} Contrastive learning aims to learn representations by pulling similar items (positives) closer together in the embedding space while pushing dissimilar items (negatives) further apart \citep{hadsell2006dimensionality, chen2020simple}. It has been widely used for learning visual representations and, more recently, for aligning representations in NLP, including cross-lingual alignment \citep{pan2021mconvert, wang2022crosslingual}.

\paragraph{Interpretability of LLM Layers} Research into LLM interpretability suggests that different layers specialize in capturing different types of information. Early layers often handle more surface-level or local features, while deeper layers capture more abstract, contextual, or task-specific information \citep{tenney2019bert, rogers2020primer}. Some studies have explored how linguistic properties, including cross-lingual information, are represented across layers \citep{lauscher2020zero}, but the specific phenomenon of strong early-layer lexical alignment versus weaker output alignment for LRLs, and targeted interventions based on this, remains less explored.

Our TLI approach distinguishes itself by: (a) empirically identifying an optimal internal model layer where pre-existing cross-lingual lexical alignment is strongest; (b) using LoRA for parameter-efficient fine-tuning; and (c) applying a contrastive learning objective directly at this identified optimal internal layer. This strategy aims to unlock and amplify the model's latent capabilities rather than solely relying on surface-level data or teaching alignment from scratch at the output layer.

\section{Methodology} \label{sec:methodology}
Our methodology comprises three main stages: (1) identifying the optimal pre-existing alignment layer within the base LLM via a pilot study; (2) Word pair curation for training and evaluation; and (3) implementing and evaluating Targeted Lexical Injection (TLI) using LoRA fine-tuning.

\subsection{Base Model}
We use Lugha-Llama-8B-wura \citep{LughaLlama2023} as our base model. Lugha-Llama is an open-source LLM specifically adapted for several African languages, including Swahili, built upon the Llama-3 architecture. We utilize the 8-billion parameter "wura" variant. The model is loaded in 4-bit precision using bitsandbytes \citep{dettmers2022llmint8} with NF4 quantization and \texttt{torch.bfloat16} as the compute data type to manage computational resources effectively.

\subsection{Identifying the Optimal Pre-existing Alignment Layer (Pilot Study)} \label{sec:pilot_study}
To inform our TLI strategy, we first conducted a pilot study to determine which layer of the Lugha-Llama-8B-wura model exhibits the highest degree of inherent cross-lingual lexical alignment for Swahili-English word pairs before any fine-tuning.

\paragraph{Procedure}
A curated set of Swahili-English word pairs (described in Section \ref{sec:word_pair_curation}) was used for this initial scan.
For each word in a pair, its embedding was extracted from the output of every transformer layer in Lugha-Llama (Layers 0 through 31, where Layer 0 represents the initial input embeddings).
Embeddings were generated by tokenizing the word, passing it through the model, and extracting the hidden states from the specified layer. Mean pooling over the attention-masked token embeddings was employed to obtain a single vector representation for each word.
Crucially, both the Swahili and English word embeddings from each layer were L2-normalized.
The cosine similarity between the L2-normalized Swahili embedding and its corresponding L2-normalized English embedding was calculated for each pair at each layer.
The average cosine similarity across all word pairs was then computed for each layer.

\begin{figure}[htbp]
    \centering
    \includegraphics[width=0.8\textwidth]{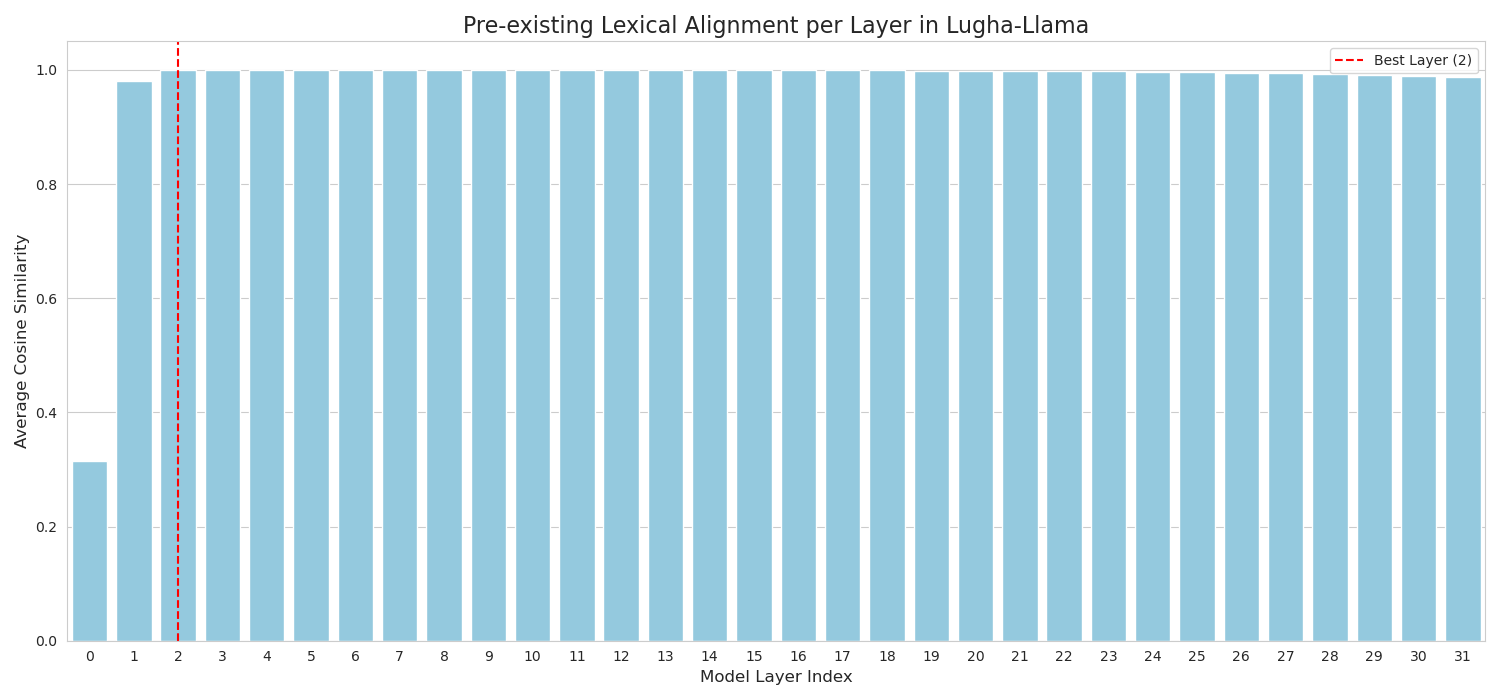} 
    \caption{Average cosine similarity between Swahili-English translation pairs across all 32 transformer layers of the Lugha-Llama-8B-wura model (pre-TLI) during the pilot study. Layer 0 represents input embeddings. Layer 2 (indicated by the red dashed line) exhibits the peak pre-existing lexical alignment ($\sim$0.99998 average similarity) and was selected as the TARGET\_LAYER for TLI.}
    \label{fig:layer_similarity_pilot}
\end{figure}

\paragraph{Results of Pilot Study}
The analysis, visually summarized in Figure \ref{fig:layer_similarity_pilot}, revealed a striking pattern.
Layer 0 (input embeddings) showed a modest average cosine similarity of approximately 0.3153.
A dramatic increase was observed at Layer 1, with an average similarity of 0.9808.
Layer 2 exhibited the peak average cosine similarity, reaching 0.99998, indicating near-perfect alignment within this specific layer during this scan. This remarkable alignment, differing from perfect unity potentially due to floating-point precision or minimal processing noise, suggests the model has already performed significant lexical mapping at this early stage.
Subsequent layers (3-31) maintained very high similarity in this pilot scan, though with a slight, gradual decrease, with Layer 31 showing an average similarity of 0.9876. It is important to note that this high similarity at Layer 31 in the pilot scan may reflect the specific conditions or word subset used for this comprehensive layer-wise analysis, and differs from the baseline output similarity ($\sim$0.32) observed on our full evaluation set (see Section \ref{sec:results_preexisting}), which represents the performance TLI aims to improve.

\paragraph{Conclusion of Pilot Study} This pilot study empirically demonstrated that Lugha-Llama-8B-wura possesses a strong, almost perfect inherent lexical alignment capability for Swahili-English pairs at Layer 2. This layer was therefore selected as the TARGET\_LAYER for extracting embeddings during the TLI fine-tuning process.

\subsection{Word Pair Curation} \label{sec:word_pair_curation}
A dataset of 686 unique Swahili-English word pairs was meticulously curated. These pairs were primarily sourced from the SAWA Corpus \citep{sim2022sawa}, focusing on common vocabulary, culturally relevant terms, and expressions that might pose challenges for code-switching or translation. The curated set includes both single words and common multi-word expressions (MWEs). This dataset was then split into:
\begin{itemize}
    \item \textbf{Trained Set:} 623 pairs used for the TLI fine-tuning process.
    \item \textbf{Control Set:} 63 pairs, semantically distinct from the trained set (ensuring no direct semantic overlap or close morphological relation beyond general language characteristics) and entirely unseen during TLI fine-tuning, used to evaluate the generalization capabilities of the TLI method.
\end{itemize}

\subsection{Targeted Lexical Injection (TLI) via LoRA Fine-Tuning}
The TLI process aims to fine-tune the Lugha-Llama model using LoRA adapters such that the strong lexical alignment observed internally at Layer 2 is better preserved and reflected in the model's final output representations.

\paragraph{Embedding Extraction for TLI Training}
During TLI fine-tuning, for each Swahili anchor word and its corresponding English positive translation in the training batch, their respective embeddings were extracted from the output of TARGET\_LAYER = 2 of the Lugha-Llama model. Consistent with the pilot study, these embeddings were mean-pooled over attention-masked tokens and then L2-normalized.

\paragraph{LoRA Configuration}
We employed LoRA with the following configuration:
\begin{itemize}
    \item Rank (r): 16
    \item Alpha (lora\_alpha): 32
    \item Dropout (lora\_dropout): 0.05
    \item Target Modules: ["q\_proj", "v\_proj"] (query and value projection matrices in the self-attention mechanisms)
    \item Task Type: CAUSAL\_LM
    \item Bias: none
\end{itemize}

\paragraph{Contrastive Loss Function}
A contrastive learning objective based on a triplet margin loss was used. For an anchor Swahili embedding ($e_{aL2}$ from Layer 2) and its positive English translation embedding ($e_{pL2}$ from Layer 2), we aim to minimize their distance while maximizing the distance to negative English embeddings. The loss for a given triplet is:
\begin{equation}
L = \max(0, \text{margin} + \text{sim}(e_{aL2}, e_{nL2}) - \text{sim}(e_{aL2}, e_{pL2}))
\end{equation}
where sim denotes cosine similarity, and margin is a hyperparameter set to 0.4 (a commonly used value in similar contrastive learning setups).

\paragraph{Negative Sampling} We utilized an in-batch negative sampling strategy. For each anchor $e_{aL2}$ in a batch, all other positive English embeddings $e_{pjL2}$ (where $j \neq i$) in the same batch served as potential negative candidates. The "hardest" negative $e_{nL2}$ -- the one with the highest cosine similarity to $e_{aL2}$ among these candidates -- was selected for the loss calculation. This is implemented in the \texttt{contrastive\_loss\_in\_batch\_negatives\_vectorized} function detailed in our training script (see Appendix \ref{app:script_details}).

\paragraph{Training Details}
\begin{itemize}
    \item Optimizer: AdamW \citep{loshchilov2017decoupled}
    \item Learning Rate: 2e-4
    \item Epochs: 5
    \item Batch Size: 8
    \item Warmup Steps: 50 (using a linear learning rate scheduler with warmup)
\end{itemize}
The model was trained on a device equipped with CUDA.

\subsection{Evaluation Protocol}
To assess the impact of TLI, we compared the lexical alignment performance of the base Lugha-Llama-8B-wura model (Pre-TLI) against the same model with the trained TLI LoRA adapters merged (Post-TLI).

\paragraph{Embedding Extraction for Evaluation}
For this evaluation phase, word embeddings for both Swahili and English words were extracted from the final output layer (Layer 31, equivalent to the model's last hidden state output) of both the Pre-TLI and Post-TLI models. This ensures we measure the alignment as reflected in the representations the model would typically use for downstream tasks. Embeddings were mean-pooled over attention-masked tokens and L2-normalized.

\paragraph{Metric}
Cosine similarity between the L2-normalized Swahili and English word embeddings was used as the primary metric for lexical alignment.

\paragraph{Evaluation Sets}
Performance was measured on both the Trained Set (623 pairs) and the Control Set (63 unseen pairs).

\paragraph{Statistical Analysis}
A paired t-test was conducted to determine the statistical significance of the observed changes in mean cosine similarity before and after TLI for both evaluation sets.

\section{Results} \label{sec:results}
This section presents the results of our experiments, starting with a recap of the pre-existing alignment and then detailing the impact of TLI.

\subsection{Pre-existing Lexical Alignment in Lugha-Llama (Recap)} \label{sec:results_preexisting}
As detailed in Section \ref{sec:pilot_study} and illustrated in Figure \ref{fig:layer_similarity_pilot}, the pilot study revealed that Lugha-Llama-8B-wura inherently achieves very high lexical alignment in its early layers, peaking at Layer 2 with an average cosine similarity of 0.99998 under the pilot study conditions. This contrasts with the significantly lower average similarity of approximately 0.3211 (for the trained set) and 0.3143 (for the control set) observed at the final output layer (Layer 31) of the base model when evaluated on our full curated word pair sets prior to TLI fine-tuning. This disparity underscores that while the model possesses strong latent alignment capabilities early on, this is not effectively propagated to its final output representations for the task at hand.

\subsection{Impact of TLI on Lexical Alignment (Quantitative Results)}
The TLI fine-tuning, targeting Layer 2 embeddings, aimed to improve the propagation of this strong internal alignment to the model's final output layer. Table \ref{tab:tli_impact} summarizes the quantitative results on both the trained and control word pair sets, comparing mean cosine similarity at the final output layer before and after TLI.

\begin{table}[htbp]
\centering
\caption{Impact of TLI on mean cosine similarity of Swahili-English word pairs at the final output layer.}
\label{tab:tli_impact}
\resizebox{\textwidth}{!}{%
\begin{tabular}{@{}lccccccr@{}}
\toprule
Evaluation Set & N & Pre-TLI Mean Sim. (Std.Dev.) & Post-TLI Mean Sim. (Std.Dev.) & Abs. Impr. & \% Impr. & T-statistic & p-value \\
\midrule
Trained Pairs & 623 & 0.3211 (0.0834) & 0.4113 (0.0877) & +0.0902 & +28.08\% & 54.8842 & $< 1.33 \times 10^{-240}$ \\
Control (Unseen) Pairs & 63 & 0.3143 (0.0734) & 0.4033 (0.0788) & +0.0890 & +28.32\% & 18.4525 & $< 7.17 \times 10^{-27}$ \\
\bottomrule
\end{tabular}%
}
\end{table}

The results clearly demonstrate a substantial and statistically significant improvement in lexical alignment due to TLI:
\begin{itemize}
    \item \textbf{Trained Word Pairs:} The mean cosine similarity for the 623 pairs used during TLI training increased from 0.3211 to 0.4113, an absolute improvement of +0.0902 and a relative improvement of +28.08\%. The paired t-test yielded a t-statistic of 54.8842 and an extremely small p-value ($< 1.33 \times 10^{-240}$), indicating that this improvement is overwhelmingly statistically significant.
    \item \textbf{Control (Unseen) Word Pairs:} Remarkably, the 63 control word pairs, which were not seen during TLI training, exhibited a comparable improvement. Their mean cosine similarity increased from 0.3143 to 0.4033, an absolute improvement of +0.0890 and a relative improvement of +28.32\%. This improvement is also highly statistically significant (t-statistic: 18.4525, p-value: $< 7.17 \times 10^{-27}$).
\end{itemize}
These results strongly support both our hypotheses: TLI effectively enhances output-level lexical alignment for targeted vocabulary, and this enhancement robustly generalizes to unseen vocabulary.

\subsection{Qualitative Examples}
To illustrate the impact of TLI, Table \ref{tab:qualitative_examples} presents examples of Swahili-English word pairs from our dataset (drawn from the 'trained' set), showing their cosine similarity at the final output layer before and after TLI.

\begin{table}[htbp]
\centering
\caption{Qualitative examples of cosine similarity changes post-TLI.}
\label{tab:qualitative_examples}
\begin{tabular}{@{}llccr@{}}
\toprule
Swahili & English & Pre-TLI Sim. & Post-TLI Sim. & Change \\
\midrule
asante sana & thank you very much & 0.292 & 0.440 & +0.148 \\
bei gani & how much does it cost & 0.309 & 0.484 & +0.174 \\
Kamba & rope & 0.326 & 0.409 & +0.083 \\
afya & health & 0.280 & 0.364 & +0.084 \\
-chafuka & rough & 0.220 & 0.326 & +0.106 \\
chungwa & orange & 0.325 & 0.335 & +0.010 \\
Februari & February & 0.525 & 0.519 & -0.006 \\
dada & sister & 0.546 & 0.541 & -0.005 \\
\bottomrule
\end{tabular}
\end{table}

The examples show substantial gains for many pairs, including single words (e.g., 'Kamba', 'afya') and multi-word expressions (e.g., 'asante sana', 'bei gani'). As expected within a large dataset, a very small number of pairs (e.g., 'Februari', 'dada' in this selection) showed marginal or no improvement, or a slight decrease. This does not detract from the overwhelmingly positive average improvement and strong statistical significance observed across the entire dataset.

\subsection{Visualization of Embedding Space}
\begin{figure}[htbp]
    \centering
    \includegraphics[width=0.8\textwidth]{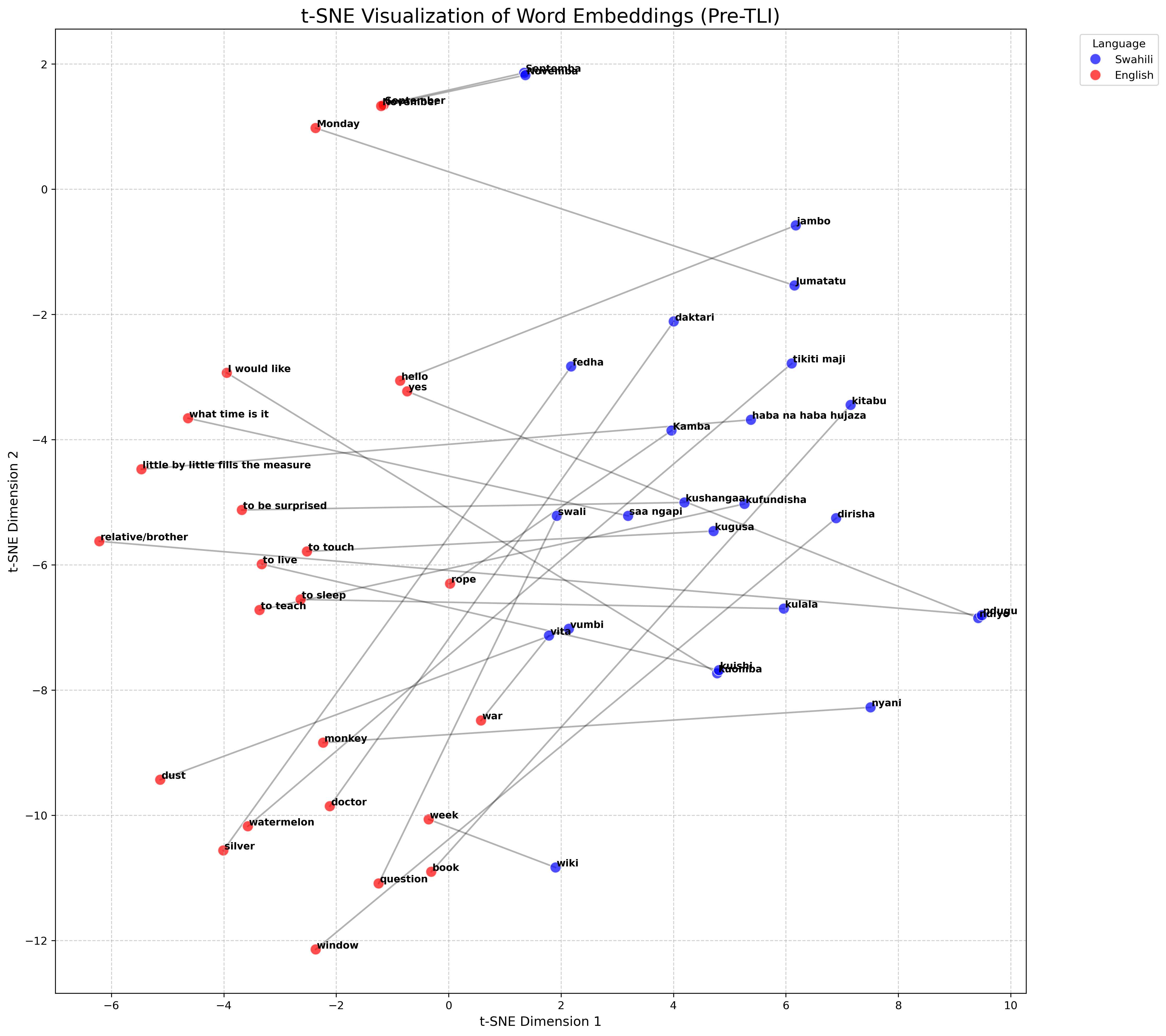} 
    \caption{t-SNE visualization of final layer Swahili (blue) and English (red) word embeddings before TLI. Lines connect translation equivalents. Many pairs are distant, indicating weaker output-level alignment.}
    \label{fig:tsne_pre_tli}
\end{figure}

\begin{figure}[htbp]
    \centering
    \includegraphics[width=0.8\textwidth]{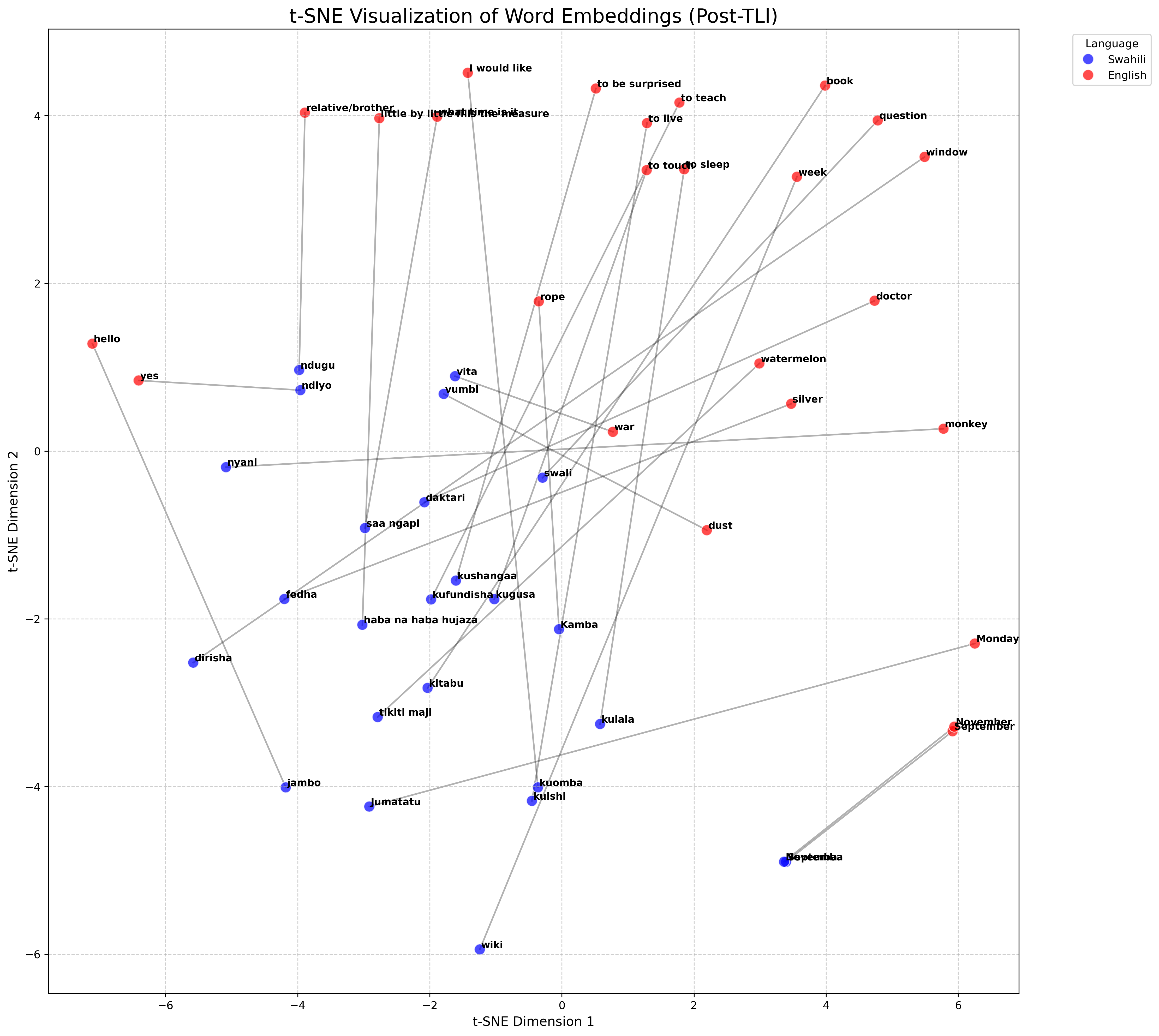} 
    \caption{t-SNE visualization of final layer Swahili (blue) and English (red) word embeddings after TLI. Lines connect translation equivalents. A noticeable improvement in alignment is visible, with corresponding pairs clustering more closely.}
    \label{fig:tsne_post_tli}
\end{figure}

To visually assess the impact of TLI on the embedding space, t-SNE visualizations of the final layer embeddings for a representative subset of Swahili-English word pairs were generated both before fine-tuning (Pre-TLI, Figure \ref{fig:tsne_pre_tli}) and after fine-tuning (Post-TLI, Figure \ref{fig:tsne_post_tli}). In these plots, Swahili words are represented by blue markers and English words by red markers. Corresponding words from translation pairs are connected by a grey line, illustrating their proximity in the 2D projected space.

Figure \ref{fig:tsne_pre_tli} (Pre-TLI) shows the initial state of the embedding space. While some related concepts might be loosely grouped, many translation pairs (e.g., 'kitabu' (book), 'dirisha' (window), 'daktari' (doctor)) are relatively distant, and the overall structure appears diffuse for many pairs. The connecting lines are often long, indicating poorer alignment at the output layer.

In contrast, Figure \ref{fig:tsne_post_tli} (Post-TLI) demonstrates a marked improvement in lexical alignment. Corresponding Swahili and English embeddings for numerous pairs (e.g., 'kitabu'/'book', 'kamba'/'rope', 'daktari'/'doctor') are visibly much closer together, with shorter connecting lines. The clusters of translation pairs appear more coherent and tightly grouped. This qualitative visual evidence, covering a sample of words (including examples from both trained and control sets), corroborates the quantitative improvements reported in Table \ref{tab:tli_impact} and suggests that TLI effectively reorganizes the final layer embedding space to better reflect translational equivalence.

\section{Discussion} \label{sec:discussion}
The experimental results provide strong evidence for the efficacy of our Targeted Lexical Injection approach and offer valuable insights into the internal workings of Lugha-Llama.

\subsection{Confirmation of Hypotheses}
Our primary hypothesis -- that TLI targeting Layer 2 would significantly improve output-level lexical alignment for trained pairs -- is unequivocally supported by the +28.08\% increase in mean cosine similarity and the extremely low p-value. Our secondary, more profound hypothesis -- that these improvements would generalize to unseen vocabulary -- is also strongly confirmed by the comparable +28.32\% improvement in the control set.

\subsection{Efficacy of TLI for Targeted Pairs}
The substantial improvement in alignment for the 623 trained word pairs demonstrates that the TLI process, involving contrastive loss on Layer 2 embeddings and LoRA updates, effectively modifies the model to better reflect these specific lexical equivalences in its final output representations.

\subsection{The Generalization Phenomenon: Unlocking Latent Capabilities}
The most significant finding of this study is the robust generalization of TLI's benefits to the unseen control word pairs. This suggests that TLI is not merely "memorizing" the trained pairs. Instead, by focusing the fine-tuning process on Layer 2 -- where the model already exhibits near-perfect intrinsic alignment -- TLI appears to be refining the model's overall mechanism for preserving and propagating this inherent cross-lingual knowledge through its deeper layers to the final output.

If Lugha-Llama already "knows" at Layer 2 that "mkate" and "bread" are equivalent, TLI trains the LoRA adapters throughout the network to ensure this understanding isn't lost or diluted by subsequent processing. This improved information flow pathway then naturally benefits any Swahili-English word pair that is strongly aligned at Layer 2, irrespective of its inclusion in the TLI training set. This interpretation explains why the control group saw improvements nearly identical to the trained group.

\subsection{The Strategic Role of Targeting Layer 2}
The pilot study's identification of Layer 2 as the locus of maximal pre-existing alignment was crucial. Targeting this layer with TLI allows the fine-tuning to reinforce an already strong and correct signal, rather than attempting to create alignment from scratch at a layer where it might be weaker or more entangled with other contextual information (like the final output layer). This makes the TLI process more of an optimization of information flow for existing knowledge, rather than de novo learning of lexical pairs.

\subsection{Implications for Low-Resource Language LLMs}
The TLI approach, particularly its generalization capability, has important implications:
\begin{itemize}
    \item \textbf{Efficiency:} It suggests that significant improvements in lexical alignment for LRLs can be achieved by fine-tuning on a relatively small, carefully curated set of word pairs, provided the intervention targets the right internal mechanisms. This is more data-efficient than methods requiring massive parallel corpora.
    \item \textbf{Parameter Efficiency:} The use of LoRA ensures that these improvements are achieved with minimal additional trainable parameters.
    \item \textbf{Improved Downstream Performance:} Enhanced lexical alignment is foundational for better performance in downstream tasks such as machine translation, cross-lingual information retrieval, and multilingual question answering.
    \item \textbf{Better Code-Switching Handling:} A model that better understands lexical equivalences is likely to handle code-switched text more coherently.
\end{itemize}

\subsection{Limitations}
This study has several limitations:
\begin{itemize}
    \item \textbf{Model and Language Pair Specificity:} The results are specific to Lugha-Llama-8B-wura and the Swahili-English language pair. While the principles may generalize, further research is needed for other models and languages.
    \item \textbf{Dataset Size:} While the generalization is promising, the training and control sets are of finite size.
    \item \textbf{Focus on Lexical Alignment:} Our evaluation focuses on isolated lexical pair similarity. The impact on broader semantic understanding, syntactic processing, or generative capabilities requires further investigation.
    \item \textbf{Interpretability:} While we hypothesize about improved information flow, the precise neural mechanisms modified by LoRA to achieve this preservation are complex and warrant deeper interpretability studies.
\end{itemize}

\section{Conclusion} \label{sec:conclusion}
This paper introduced Targeted Lexical Injection (TLI), a LoRA-based fine-tuning method designed to enhance cross-lingual lexical alignment in LLMs by leveraging their inherent, but often underutilized, internal capabilities. Our pilot study on Lugha-Llama-8B-wura revealed that strong Swahili-English lexical alignment (average cosine similarity $\sim$0.99998) exists in its early internal layers (specifically Layer 2), contrasting with weaker alignment ($\sim$0.32) at its output interface on our evaluation set.

By targeting Layer 2 embeddings with a contrastive learning objective, TLI significantly improved the output-level mean cosine similarity for 623 trained Swahili-English word pairs by +28.08\% (from 0.3211 to 0.4113). Crucially, this improvement generalized robustly to 63 unseen control pairs, which saw a +28.32\% increase (from 0.3143 to 0.4033). Both improvements were highly statistically significant.

These findings indicate that TLI effectively helps the model preserve and propagate its strong, latent early-layer cross-lingual knowledge to its final output. This approach offers a parameter-efficient and data-efficient pathway to improving lexical understanding in LLMs for low-resource languages, with benefits extending beyond the explicitly trained vocabulary. TLI underscores the potential of targeted interventions that work in concert with a model's existing internal knowledge structures.

\section{Future Work}
Future research will explore several avenues:
\begin{itemize}
    \item \textbf{Mechanism of Generalization:} Conduct more granular analyses to understand which types of unseen words (e.g., based on semantic categories, frequency, morphological similarity to trained words) benefit most from TLI's generalization. Investigate if LoRA is learning to preserve embedding "angles" or down-weight disruptive transformations in deeper layers.
    \item \textbf{Optimal Layer Exploration:} Investigate whether Layer 2 is universally optimal or if the ideal target layer varies across different LLM architectures, language pairs, or specific alignment tasks.
    \item \textbf{Downstream Task Evaluation:} Quantify the impact of TLI-induced lexical alignment improvements on concrete downstream NLP tasks such as Swahili-English machine translation, cross-lingual question answering, and sentiment analysis.
    \item \textbf{Active Learning for Pair Selection:} Develop active learning strategies to select the most informative word pairs for TLI training to maximize efficiency and impact.
    \item \textbf{Extension to Morphologically Rich Languages:} Adapt and evaluate TLI for languages with more complex morphology than Swahili, where tokenization and sub-word representations play a more significant role.
    \item \textbf{Broader Language Coverage:} Apply and evaluate TLI across a wider range of low-resource language pairs.
\end{itemize}

\section*{Appendix} 
\addcontentsline{toc}{section}{Appendix} 
\subsection{Training Script Details} \label{app:script_details}
Details regarding the \texttt{contrastive\_loss\_in\_batch\_negatives\_vectorized} function and other training script specifics would typically be provided here. For the purpose of this document, it refers to the conceptual implementation as described in the methodology.

\end{document}